\documentclass[10pt,twocolumn,letterpaper]{article}

\usepackage[pagenumbers]{wacv} 

\usepackage{graphicx}
\usepackage{amsmath}
\usepackage{amssymb}
\usepackage{booktabs}

\usepackage[pagebackref,breaklinks,colorlinks]{hyperref}

\usepackage[capitalize]{cleveref}
\crefname{section}{Sec.}{Secs.}
\Crefname{section}{Section}{Sections}
\Crefname{table}{Table}{Tables}
\crefname{table}{Tab.}{Tabs.}

\begin{document}

\title{AI-Driven Innovations in Volumetric Video Streaming: A Review}

\author{Erfan Entezami\\
University of Massachusetts Amherst\\
Amherst, MA, USA\\
{\tt\small eentezami@cs.umass.edu}
\and
Hui Guan\\
University of Massachusetts Amherst\\
Amherst, MA, USA\\
{\tt\small huiguan@cs.umass.edu}
}
\maketitle

\begin{abstract}
Recent efforts to enhance immersive and interactive user experiences have driven the development of volumetric video, a form of 3D content that enables 6 DoF. Unlike traditional 2D content, volumetric content can be represented in various ways, such as point clouds, meshes, or neural representations. However, due to its complex structure and large amounts of data size, deploying this new form of 3D data presents significant challenges in transmission and rendering. These challenges have hindered the widespread adoption of volumetric video in daily applications. 
In recent years, researchers have proposed various AI-driven techniques to address these challenges and improve the efficiency and quality of volumetric content streaming. This paper provides a comprehensive overview of recent advances in AI-driven approaches to facilitate volumetric content streaming. 
Through this review, we aim to offer insights into the current state-of-the-art and suggest potential future directions for advancing the deployment of volumetric video streaming in real-world applications.
    
\end{abstract}


\section{Introduction}

Recent advances in computer vision and graphics have made it possible to move from interacting with traditional 2D content to immersive experiences that more naturally simulate real environments for users. This technological shift began with early versions of 3D video capturing \cite{matsuyama20123d,wurmlin20023d,smolic20113d,waschbusch2005scalable} and has since progressed to more advanced forms, such as 360° video displayed on Head-mounted Display (HMD)\cite{yaqoob2020survey,xu2020state,vskola2020virtual,shafi2020360}. 360° content allows users to interact with the environment by having rotational movement in three directions (roll, yaw, pitch), providing a 3-degrees-of-freedom (DoF) experience. This format has gained widespread adoption in various domains such as entertainment\cite{hebbel2017360,alqahtani2017environments,guttentag2010virtual}, healthcare \cite{minns2019immersive, kyaw2019virtual, bi2023enhancing} and education \cite{kaliraj2021innovating, bi2023enhancing,roche2017using} due to its ability to offer a more realistic experience. 

The advancement of 3D content towards a more realistic experience has progressed even further with the introduction of volumetric content. Volumetric content is a form of 3D content that provides 6 DoF for users, including 3 positional (x, y, z) and 3 rotational. 
Volumetric content gives users more flexibility to view the 3D scene from different angles and provides a more interactive experience. 
According to recent research~\cite{lee2020groot, li2022optimal}, the strong potential for using volumetric content in various domains, along with anticipated technical advances in the near future, position volumetric video as a crucial component of future video streaming systems and a key use case for 6G network~\cite{li2022optimal}.

Volumetric video can be represented using various methods, such as point clouds~\cite{jin2023capture}, neural networks~\cite{mildenhall2021nerf}, 3D Gaussian splatting (3DGS)~\cite{kerbl20233d}, and more. 
The evolution of volumetric content representation began with simpler data structures such as point clouds and meshes and over time, other representation techniques have emerged to enhance visual quality, transmission efficiency, and rendering latency. 

While deploying volumetric video offers significant benefits, there are challenges for each representation technique that prevent its widespread adoption in today’s video streaming and communication platforms. 
These challenges are related to different parts of the streaming pipelines, including capturing and creating volumetric data, storing large-size volumetric content, compression and decoding, data transmission, model training, and rendering. Researchers have been developing various solutions for these challenges across different volumetric content representation techniques to enable seamless streaming and establish volumetric content as a reliable and practical format for streaming platforms.

\begin{figure}[ht]
    \centering
    \includegraphics[width=0.47\textwidth]{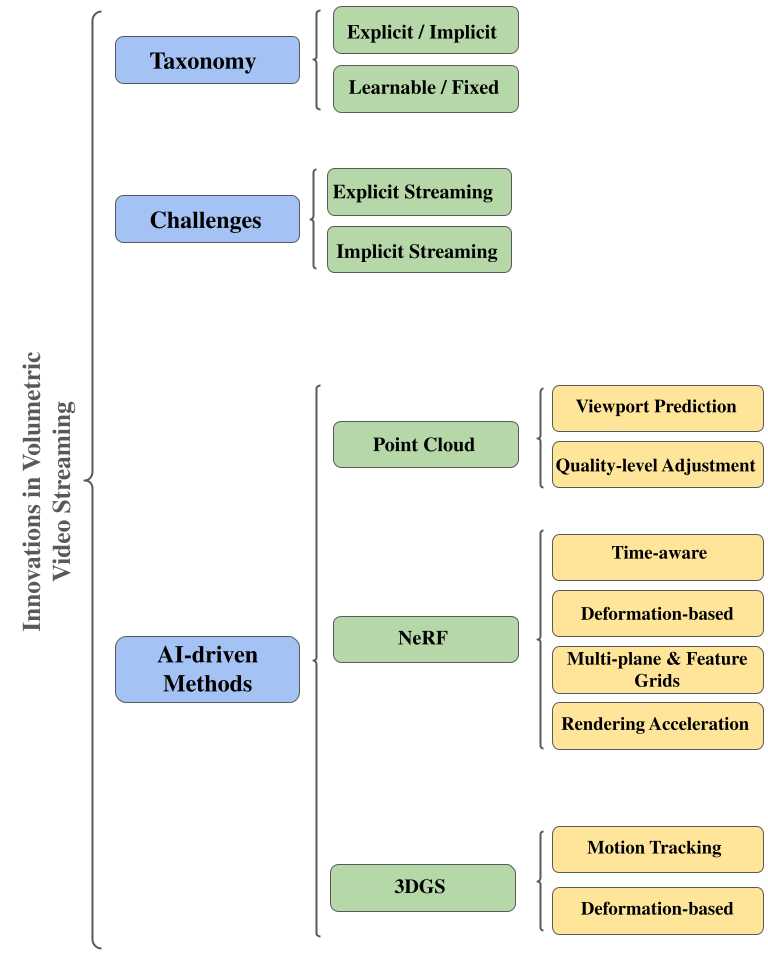} 
    \caption{The organization of this review.}
    \label{fig:organization}
\end{figure}

In this paper, we provide a summary of the challenges and solutions related to streaming volumetric video from the server to the end user such as transmitting data and rendering the model, leaving other topics, such as capturing volumetric content and storing large data, for future works.
Compared to previous surveys that have dived into technical explanations of volumetric content representations such as point cloud~\cite{smolic20113d,alkhalili2020survey,viola2023volumetric}, Neural Radiance Field (NeRF)~\cite{lin2024dynamic,gao2022nerf,rabby2023beyondpixels}, and 3DGS~\cite{wu2024recent,fei20243d,chen2024survey}, our focus is only on the AI-driven approaches developed to enable efficient streaming of volumetric content for each representation type. In the following sections, 
we first present a taxonomy of volumetric content representations in \S\ref{sec:taxonomy} and highlight the specific challenges associated with each in \S\ref{sec:challenges}. 
We then categorize and discuss the key solutions proposed to overcome these challenges in \S\ref{sec:solutions}. In \S\ref{sec:discussion}, we discuss the remaining challenges and propose several promising directions for future research and development. Finally, in \S\ref{sec:conclusion}, we conclude by summarizing our findings from the review of recent literature in the field.
Through this review, we aim to offer insights into the current state-of-the-art and suggest potential future directions for advancing the deployment of volumetric video streaming in real-world applications. The structure of our paper is shown in Figure \ref{fig:organization}.


\begin{figure}[ht]
    \centering
    \includegraphics[width=0.47\textwidth]{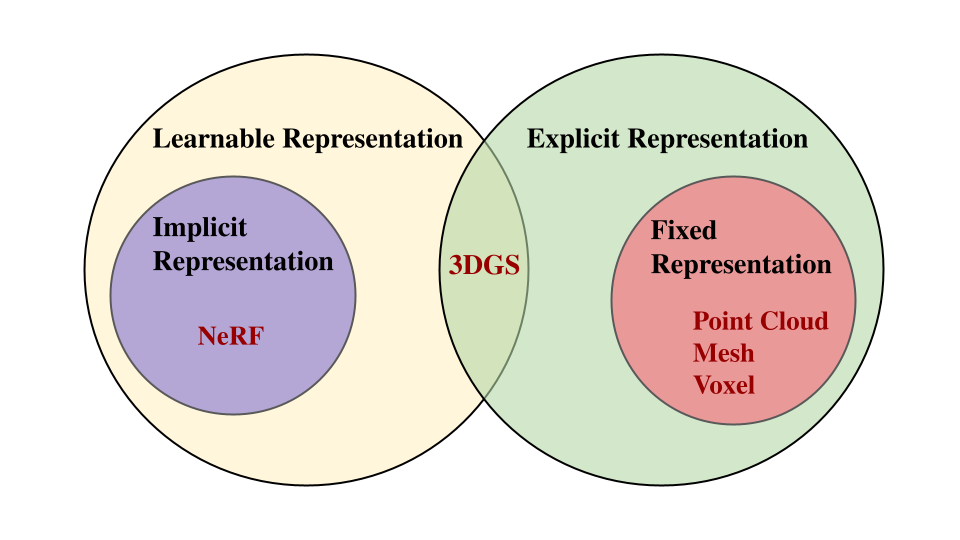} 
    \caption{Visualization of volumetric content representations, categorized as implicit or explicit, and learnable or fixed.}
    \label{fig:classes}
\end{figure}

\section{Taxonomy of Representation Techniques}
\label{sec:taxonomy}
Unlike 2D videos whose representation has reached a level of maturity, volumetric videos can be represented using various techniques, each with its own advantages and disadvantages, and each requiring different strategies for designing the streaming pipeline. In this section, we present two approaches for classifying methods for representing volumetric content and highlight the three most commonly used representation techniques namely point cloud, NeRF and 3DGS. Figure~\ref{fig:classes} illustrates the relationships between different volumetric content representations considered in this work.


\vspace{.1cm}\noindent\textbf{Explicit and Implicit Representations.} Existing representations can be categorized into explicit and implicit representations. Explicit representation uses a data structure that explicitly stores volumetric content as a type of 3D data. Each element within this structure contains 3D information such as 3D position and RGB color. Examples of explicit representation include point cloud, plenoptic point cloud, mesh, voxel and 3DGS. 

Implicit representation, on the other hand, represents the 3D scene without explicitly storing the geometry of the environment. It often involves a neural network that learns to represent the 3D scene based on given inputs. To generate the target view of a 3D scene, these methods typically send parameters such as location and viewing direction to a trained model, which then produces the target 2D image from the 3D scene. NeRF is the most known implicit representation for volumetric content.

\vspace{.1cm}\noindent\textbf{Learnable and Fixed Representations.}
Another way to classify different types of volumetric content representations is based on their ability to optimize their structure to better represent volumetric scenes, which can be either learnable or fixed. In the learnable category, methods typically involve a training process that utilizes multiple images from a 3D scene to learn how to represent it effectively. Learnable representations can be either implicit or explicit. The most promising methods proposed for learnable representations are NeRF and 3DGS, which represent scenes implicitly and explicitly, respectively.

Fixed representations refer to explicit representations that do not involve any form of training or optimization. As mentioned previously, this type of representation stores information about the 3D scene, such as geometry and color, in the form of a data structure, where accessing and modifying individual elements is usually possible. Once the structure of the 3D scene is stored, there is no further improvement or refinement involved, as the representation remains static.

Next, we highlight the details of three representative methods, point cloud, NeRF, and 3DGS.\\

\vspace{.1cm}\noindent\textbf{Point Cloud.} Point cloud is one of the earliest methods used for representing volumetric content. A point cloud as an explicit and fixed representation is an unsorted set of 3D points, each with a 3D position and attributes such as RGB color and intensity. Compared to other explicit formats like mesh and voxel, point cloud does not store additional information such as the connections between points, which offers several advantages, including easier creation and a simpler structure, which facilitate storage and rendering. This simple structure made point cloud a popular choice for volumetric content. However, transmitting data in point cloud format over WiFi network remains a challenge due to the large data size. For example, to stream point cloud-based video with 30 fps (as a know threshold for smooth video streaming in previous research~\cite{li2022optimal,han2020vivo,liu2023cav3}) with around 760,000 points per frame, a bandwidth of up to 2.9 Gbps may be required which indicates the streaming difficulties associated with this format~\cite{li2022optimal}.

\vspace{.1cm}\noindent\textbf{NeRF.} NeRF is an implicit and learnable representation that approximates  3D scenes using a neural network. NeRF-based methods typically train a Multi-Layer Perceptron (MLP) model to predict the color and density of a point given its viewing direction and position. This can be expressed as follows:
\begin{equation}
F(P, \theta, \phi) \rightarrow (c, \sigma),
\end{equation}
where $P=(x, y, z)$ is the coordinate of the point, $\theta$ and $\phi$ are polar and azimuthal angles which together represent the viewing direction in spherical coordinate system. $c = (r,g,b)$ and $\sigma$ represent predicted color and density respectively.

\begin{figure}[h]
    \centering
    \includegraphics[width=0.47\textwidth]{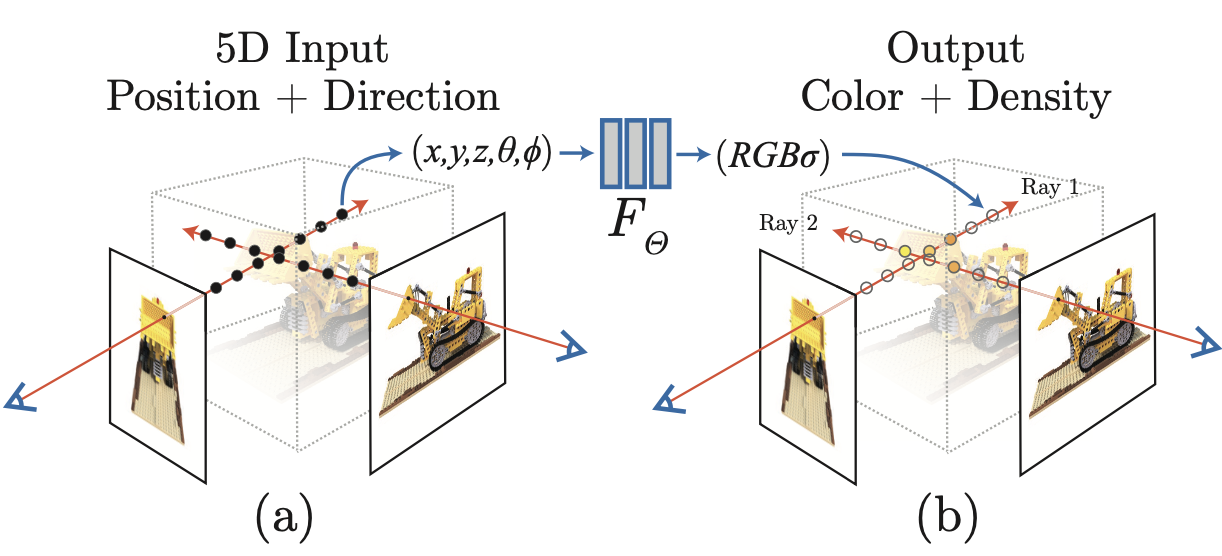} 
    \caption{Main pipeline of NeRF. In step (a) position and direction of points are sent to the MLP model to predict color and density, and  in (b) those predicted values are integrated using volume rendering techniques to achieve the final color. The image is taken from \cite{mildenhall2021nerf}.}
    \label{fig:nerf}
\end{figure}


During the rendering phase of NeRF, the trained neural radiance field is used to generate novel views of the 3D scene. To produce a 2D image corresponding to a given camera position and viewing direction, a technique called ray marching is employed. In this technique, a ray is cast from the camera's position into the scene for each pixel in the 2D image plane(of size $H \times W$), resulting in a total of $H \times W$ rays, one for each pixel in the pixel grid. Along each ray, multiple sample points are taken, and for each sample point, the color and density are predicted by querying an MLP model. These predicted colors and densities are then integrated along the ray using classical volume rendering techniques \cite{kajiya1984ray,mildenhall2021nerf,porter1984compositing} to compute the final color for each pixel. This process is repeated for all pixels in the image to generate a 2D view of the 3D scene from the given position and viewing direction.

During the training phase of NeRF, a neural network model is trained to generate a 2D image for any given viewing direction. The model uses a set of 2D images captured from various viewing angles of a 3D scene, along with corresponding camera parameters (position and viewing direction), as training data. These input data allow the model to learn the 3D scene's geometry and appearance to synthesize a 2D image from a given viewing direction, based on a learned 3D scene representation.
During the training iterations, a batch of 2D images of size $H \times W$ is generated using the rendering process explained earlier, each representing a 2D projection of the 3D scene from a different viewing direction. 

By comparing the generated 2D images to the ground truth images, a photometric loss function is computed to update the weights of the MLP model in the training phase \cite{rabby2023beyondpixels,mildenhall2021nerf}.
This loss function measures the difference between the rendered and ground truth color for each pixel, as defined below:

\begin{equation}
L = \sum_{r \in R} || C_r - \hat{C}_r ||_2^2
\end{equation}

Where $L$ is the RGB reconstruction loss , $C_r$ is the ground truth color $C_r - \hat{C}_r$ is the rendered color at a given pixel, $r$ is a ray in the set of all cast rays $R$. Figure~\ref{fig:nerf} illustrates the training and rendering of a NeRF model.


\vspace{.1cm}\noindent\textbf{3DGS.} 

\begin{figure*}[h]
    \centering
    \includegraphics[width=\textwidth]{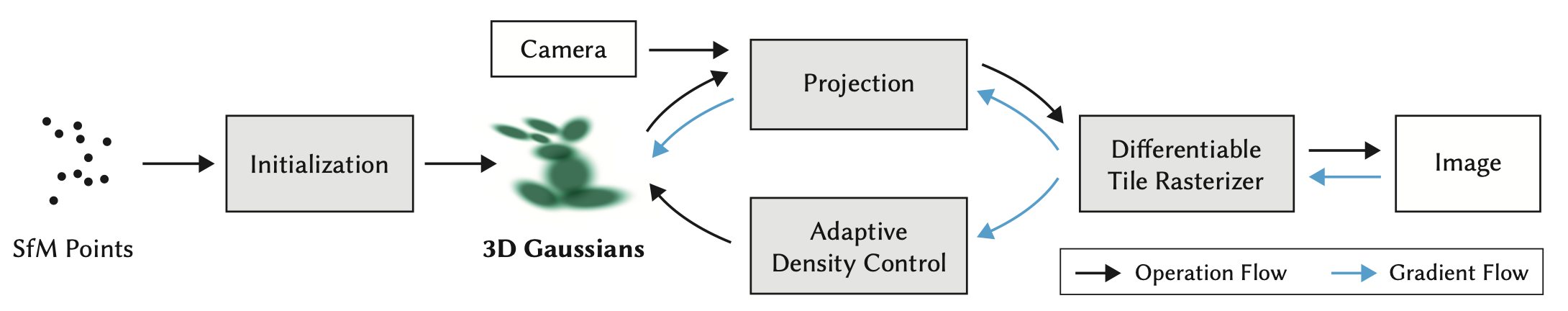} 
    \caption{The main training pipeline for 3DGS. image is taken from \cite{kerbl20233d}.}
    \label{fig:3DGS_pic}
\end{figure*}


Rendering a learned 3DGS representation involves synthesizing a 2D image from a 3D scene containing a vast number of optimized 3D Gaussians. Unlike NeRF, which relies on computationally intensive volumetric ray marching, 3DGS projects these 3D Gaussians directly onto a 2D image plane through a process called splatting. In splatting, each 3D Gaussian is projected into the 2D space as an ellipse using 3D covariance matrix $\Sigma$ and viewing transformations $W$.

After projection, 3DGS calculates the contribution of each Gaussian to each pixel which involves sorting Gaussians by depth relative to the camera and integrating their colors and opacities using alpha compositing technique \cite{porter1984compositing} to compute the final pixel values. To accelerate this process, the image plane is divided into tiles, and rendering occurs in parallel for each tile, leveraging shared memory for efficient computations. These optimizations make 3DGS suitable for real-time rendering while preserving high-quality image synthesis.

During the training phase, the parameters of the 3D Gaussians are optimized to improve the visual representation of the 3D scene. 
Initially, the 3D Gaussians are initialized based on a point cloud generated using the Structure from Motion (SfM) method \cite{snavely2006photo}, which estimates 3D structure from multiple 2D images. Similar to the training process in NeRF, these 2D images are captured from various viewing angles of the 3D scene and used as training data. However, unlike NeRF, the training images for 3DGS do not require camera parameters such as position and viewing direction. Instead, these parameters are implicitly estimated by the techniques used in SfM method. Once Gaussians are initialized, the next step involves rendering the 3DGS to simulate a 2D view of the 3D scene from a given viewing direction.
The generated 2D image is then compared to the ground truth image, and a loss function is computed based on the difference between the two images. Unlike NeRF, which calculates loss at the pixel level due to ray marching process, 3DGS focuses on image-level loss function which combines $L_1$ and $L_{\text{D-SSIM}}$ losses, which together measure the difference between the rendered and the ground truth images as follows:
\begin{equation}
L = (1 - \lambda)L_{1} + \lambda L_{\text{D-SSIM}}
\end{equation}
Where $\lambda \in [0, 1]$ is a scalar factor and $L$ denotes the 3DGS loss function.

This loss is used to compute the gradients, indicating how the parameters of the Gaussians should be adjusted. The stochastic gradient descent (SGD) algorithm leverages these gradients to optimize the learnable parameters, iteratively refining the Gaussians to improve the visual quality and accuracy of the rendered scene. Figure \ref{fig:3DGS_pic} illustrates the training pipeline for 3DGS.

\section{Challenges}
\label{sec:challenges}
This section explores the specific challenges in volumetric content streaming and clarify the motivation behind developing AI-based techniques to address these issues in subsequent sections.  
We categorize the challenges based on explicit (\ref{Challenges_explicit}) and implicit (\ref{Challenges_implicit}) representations.

\subsection{Explicit Volumetric Content Streaming}
\label{Challenges_explicit}

Using explicit volumetric content, such as point clouds, in an end-to-end streaming pipeline requires transmitting enormous amounts of data at high speeds to provide an immersive interactive experience for users. Unlike 2D videos, codecs for explicit 3D contents are not yet well-established, thus compression methods in this field often fall short, either in compression rate or in encoding and decoding speed~\cite{yang2023comparative}.

The Moving Picture Experts Group (MPEG) \cite{mpeg2020} introduced standardized compression techniques for point clouds in 2020, such as Video-based Point Cloud Compression (V-PCC)~\cite{graziosi2020overview} and Geometry-based Point Cloud Compression (G-PCC)~\cite{graziosi2020overview}, which achieve good compression rates. However, their slow encoding and decoding processes limit their practicality for end-to-end streaming systems~\cite{yang2023comparative}. For example, encoding a point cloud-based volumetric video called Longdress~\cite{dEon2017}, with a data size of only 511 MB using the V-PCC method, takes 19 minutes and decoding phase takes around 34 seconds~\cite{li2022optimal}. Google’s Draco codec \cite{draco} offers faster encoding and decoding speeds which makes it more appropriate for the real-time applications and is widely used in streaming pipelines for point cloud-based volumetric content~\cite{han2020vivo,lee2020groot,li2022optimal}. However, Draco’s compression rate is relatively low compared to other methods such as V-PCC and G-PCC.

Neural-based codecs such as PU-GCN+~\cite{qian2021pu} and MPU+~\cite{yifan2019patch} demonstrate better encoding latency and more stable performance in preserving content quality during compression. However, similar to V-PCC and G-PCC, their high decoding latency remains a barrier to their widespread application in real-time volumetric content streaming~\cite{yang2023comparative}.

Compression methods for 3DGS have been an active area of research due to its rapid adoption in the computer vision and graphics communities. 
Various strategies, including attribute pruning \cite{fan2023lightgaussian,girish2025eagles}, vector quantization\cite{bagdasarian20243dgs}, and structured representations \cite{lu2024scaffold, lee2024compact}, have been proposed for 3DGS, achieving remarkable success with over 80\% reduction in data storage \cite{huang2024hierarchical, lee2024compact, chen2025hac}. These strategies show potential when combined with other optimization techniques for volumetric content. However, for high-quality and long volumetric videos, relying solely on compression techniques may not be sufficient to guarantee a satisfactory experience for the end users.

Currently, 5G technology supports a theoretical transmission speed of 20 Gbps. However, previous studies on network indicate that in practice, the actual transmission rates are typically lower than these theoretical thresholds~\cite{riiser2013commute,van2016http}. 
Recent measurement studies showing observed speeds reaching up to 3 Gbps in 5G network~\cite{narayanan2021variegated} which indicates that 5G transmission speed may not be sufficient to deliver a fully immersive experience with high-quality volumetric content, especially in interactive domains that require low-latency in sending and receiving feedback~\cite{aijaz2016realizing,van2023tutorial}.

\subsection{Implicit Volumetric Content Streaming}
\label{Challenges_implicit}

Implicit representations reduce the amount of data and bandwidth needed for the transmission. In this approach, a trained model can generate a target view of the 3D scene on demand, and only transmit the generated view to the users. However, these methods typically require repeatedly querying a large neural network trained on the 3D scene to generate each target view in real time. This process leads to slow inference time, resulting in rendering delays which can significantly impact the user experience~\cite{gao2022nerf,yan2024neural}. The latency becomes particularly more problematic for dynamic scenes~\cite{du2021neural,li2023steernerf,liu2023toward}.

Generally, for implicit representations, there is an inherent trade-off between different factors such as model size, inference speed, memory consumption and visual quality, which must be optimized to provide a seamless, immersive experience for users~\cite{li2023steernerf, liu2023toward}.

\section{Recent Methods and Innovations}
\label{sec:solutions}
The challenges associated with each representation technique for volumetric videos highlight the need for advanced AI techniques to address them and enable smooth streaming systems for volumetric content delivery. In this work, we primarily focus on techniques related to streaming and rendering, leaving other considerations, such as memory storage and content capturing methods, for future research. 
We categorize the techniques developed by researchers for three widely used representations namely point cloud (\ref{methods_pointcloud}), NeRF (\ref{methods_nerf}), and 3DGS 
(\ref{methods_3dgs}). Within each category, we further classify the proposed techniques based on their main ideas and the general directions they have explored.

\subsection{Point cloud-based Streaming}
\label{methods_pointcloud}
Streaming volumetric content represented by point clouds presents challenges due to the large data transmission requirements over limited network bandwidth. The techniques discussed in this section focus on various strategies to reduce the amount of data transmitted over WiFi networks. These techniques can be categorized into two main classes namely viewport prediction (\ref{pointcloud_Viewport Prediction Methods}) and quality-level adjustment (\ref{pointcloud_Quality-level Adjustment Methods}) methods.

\subsubsection{Viewport Prediction Methods}
\label{pointcloud_Viewport Prediction Methods}

Viewport prediction methods \cite{han2020vivo,li2022optimal,liu2023cav3,lee2020groot,chen2024toward,hu2023understanding, li2023toward} aim to detect the portion of the video the user is likely to view in the upcoming timesteps, which allows reducing the amount of data that needs to be transmitted.
Though volumetric content allows viewers to explore a 3D frame with full 360 degree rotational freedom, previous research in this domain shows that viewers typically only watch a portion of the frame which is approximately 120 degree of the entire view~\cite{assarsson2000optimized, jin2023capture,han2020vivo}. 
The viewport refers to the portion of the video that is visible to the viewer. 

Viewport prediction methods employ different techniques to predict the user’s future viewport and, combined with a tile-based streaming strategy, ensure that only the relevant tiles are transmitted in high quality. In a tile-based strategy, the video frame is divided into smaller tile (in the case of volumetric content, they're also called cube), with each tile independently encoded, transmitted to the user and decoded at the destination. By integrating viewport prediction methods with tile-based streaming, tiles in the user’s anticipated viewport can be prioritized for high-quality transmission, while non-essential tiles can either be omitted or sent at a lower quality. This approach reduces required bandwidth for data transmission while ensuring optimal quality in the user’s viewing area.

Hu et al.\cite{hu2023understanding} explored and analyzed user behavior while watching volumetric video and provided one of the first publicly available datasets for viewport prediction. In \cite{han2020vivo}, a visibility-aware streaming method is proposed, where a viewport prediction model, trained on data collected from users, is combined with a tile-based strategy. This approach transmits tiles in the user’s predicted future viewport at higher quality, while streaming other tiles at lower quality to reduce the overall size of the transmitted data efficiently. Similar ideas are proposed in~\cite{li2022optimal,liu2023cav3}. Li et al. \cite{li2022optimal} consider using a hybrid tile size to more accurately cover the predicted viewport and effectively reduce the number of target tiles required for high-quality transmission. In addition to viewport prediction, their approach introduces a caching mechanism to further enhance streaming smoothness. In a subsequent study, Li et al.~\cite{li2023toward} consider the problem of target tiles selection as an optimization problem and leverage reinforcement learning to find the optimal policy for selecting them. 

Although viewport prediction methods effectively reduce the size of transmitted data, they often struggle to accurately predict user behavior over longer time windows. Additionally, training a viewport prediction model is challenging due to its reliance on efficiently collected user data and the complex feature extraction from volumetric scenes. These factors make such methods less practical for live streaming and difficult to adapt for new content.

\subsubsection{Quality-level Adjustment Methods}
\label{pointcloud_Quality-level Adjustment Methods}
Similar to viewport prediction, methods in this category aim to transmit the entire video or a portion of it in lower quality to reduce the amount of transmitted data. However, these methods do not rely on viewport prediction.

Super-resolution methods which previously have been used in streaming 2D videos, have been applied to volumetric video streaming~\cite{zhang2021efficient,zhang2022yuzu,zhang2020mobile}. In these methods, a lower-quality version of the video which has lower data size is transmitted to the end user. The destination then enhances the video resolution to deliver an appropriate visual quality while reducing bandwidth usage.

The concept of context-aware streaming has been also explored in previous works~\cite{zhu2022semantic,jin2023capture}, but most related methods in this group are classified under viewport prediction-based techniques since they leverage viewport prediction. In context-aware methods, the most important portions of the video, identified based on video context, are transmitted at higher quality using tile-based strategies or techniques like background removal and object detection. These methods often utilize models such as PointNet++ \cite{qi2017pointnet++} to extract features from the 3D scenes. 

\begin{figure}[h]
    \centering
    \includegraphics[width=0.47\textwidth]{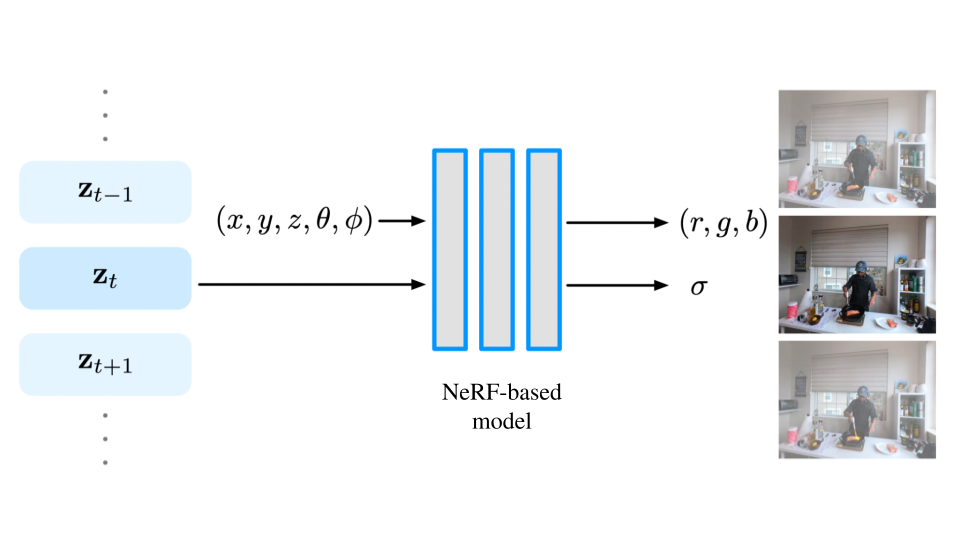} 
    \caption{An example of time-aware methods that generates a 2D view of a dynamic scene based on the position, viewing direction, and a compact time-varying latent code. Image is adapted from \cite{li2022neural}.}
    \label{fig:nerf_time_aware}
\end{figure}

\subsection{NeRF-based Streaming}
\label{methods_nerf}
Since NeRF has shown remarkable performance in representing static scenes~\cite{mildenhall2021nerf,barron2021mip, barron2022mip, tancik2022block} in terms of visual quality, researchers have explored its deployment for dynamic scenes as well. A straightforward approach to model dynamic scenes with NeRF is to either train a large model for the entire video \cite{xian2021space} or train multiple NeRF models, each corresponding to a frame. However, these approaches are often impractical, as they result in either a very large model with slow training and rendering time or an excessive number of trained models, which require substantial storage space \cite{gao2021dynamic}. 

To address these challenges, researchers have developed alternative techniques, which are reviewed in this section. We categorize and briefly explain the most common methods proposed for dynamic NeRF that can generalize to all types of volumetric content and have gained significant attention in the community. These methods can be classified into time-aware methods (\ref{nerf_Time-aware Methods}), deformation-based methods (\ref{nerf_deformation}), multi-plane and feature grids methods (\ref{nerf_Multi-plane and Feature Grids Methods}) and rendering acceleration methods (\ref{nerf_Rendering Acceleration Methods}).

\subsubsection{Time-aware Methods}
\label{nerf_Time-aware Methods}
Time-aware methods~\cite{xian2021space, li2022neural, li2021neural, gao2021dynamic} are methods that incorporate time as an additional input to the model, enabling the generation of the scene at the requested position, viewing direction and viewing time. As previously mentioned, adding time as an additional input can result in slow training and/or slower rendering. Therefore, the methods in this group have been proposed to overcome these limitations.

Methods proposed in \cite{gao2021dynamic,xian2021space} introduce strategies to create a global representation of the scene by blending a static NeRF model (to represent static components), with a time-dependent dynamic NeRF (to represent moving components), and using scene depth estimation methods respectively. Li et al. \cite{li2022neural} proposed adding additional compact latent code as an input to the model to consider time dimension. An example of time-aware methods is shown in Figure \ref{fig:nerf_time_aware}.

Time-aware methods typically require additional components, such as depth maps or extensive observations~\cite{cao2023hexplane} which may not always be feasible to provide, or depend on rendering large models, which may be slow for rendering~\cite{peng2023representing}.

\subsubsection{Deformation-based Methods}
\label{nerf_deformation}
Deformation methods were among the first approaches proposed for using NeRF for dynamic scenes~\cite{park2021nerfies, park2021hypernerf, tretschk2021non, yang2022banmo, pumarola2021d, kirschstein2023nersemble,song2023nerfplayer}. 
A deformation model can represent motion between frames. In other words, it describes how the coordinates of points are deformed or displaced between consecutive frames.

The main approach in most of deformation-based methods rely on capturing geometry of the scene using a static canonical radiance field, and training a deformation field to represent displacements of points at each timestep. 
Deformation-based methods use a mapping function like $\Psi$ that can map coordinates of each point $p=(x,y,z)$ in observation space to its corresponding coordinates $p'=(x',y',z')$ in caniconal space over time which can be formulated as $\Psi_i  : \mathbf{p} \rightarrow  \mathbf{p}' \quad \text{for} \ i \in \{1, 2, \dots, n\}$ which $n$ is number of frames.

Recently, a group of methods~\cite{liu2022devrf,fang2022fast,guo2022neural, liu2023robust, kirschstein2023nersemble} has proposed using an explicit representation such as voxel grid in combination with a deformation neural network to train NeRF-based models for dynamic scenes. 
A voxel grid is an explicit representation that stores features such as density and color within its grid cells. The use of voxel grids has shown significant improvements in training static NeRF models \cite{sun2022direct, yu2021plenoxels} and has motivated advancements in dynamic NeRFs as well. The primary advantage of using a voxel grid is that it replaces the time-consuming neural network queries with faster voxel grid interpolation. Interpolation for a point $x$ from a voxel grid $V$ can be formulated as follows:
 \begin{equation}
 \text{interp}(x, V) : (\mathbb{R}^3 \times \mathbb{R}^{C \times N_x \times N_y \times N_z}) \rightarrow \mathbb{R}^C
\end{equation}
Here, \(C\) represents the feature space, and \(N_x \times N_y \times N_z\) denotes the total number of voxels.


These methods primarily rely on extracting 3D features from an explicit representation and deformed coordinates from a deformation model, and employing shallow MLP models to obtain the final color and density for each pixel. The general formula for rendering NeRF models to obtain final density and color is similar to rendering a static NeRF model, but with the difference of having additional input parameters. Although deformation-based methods have shown significant results in various domains, their performance might decrease in scenes with large or sudden motions \cite{li2021neural,park2021nerfies}. 


\begin{figure}[h]
    \centering
    \includegraphics[width=0.47\textwidth]{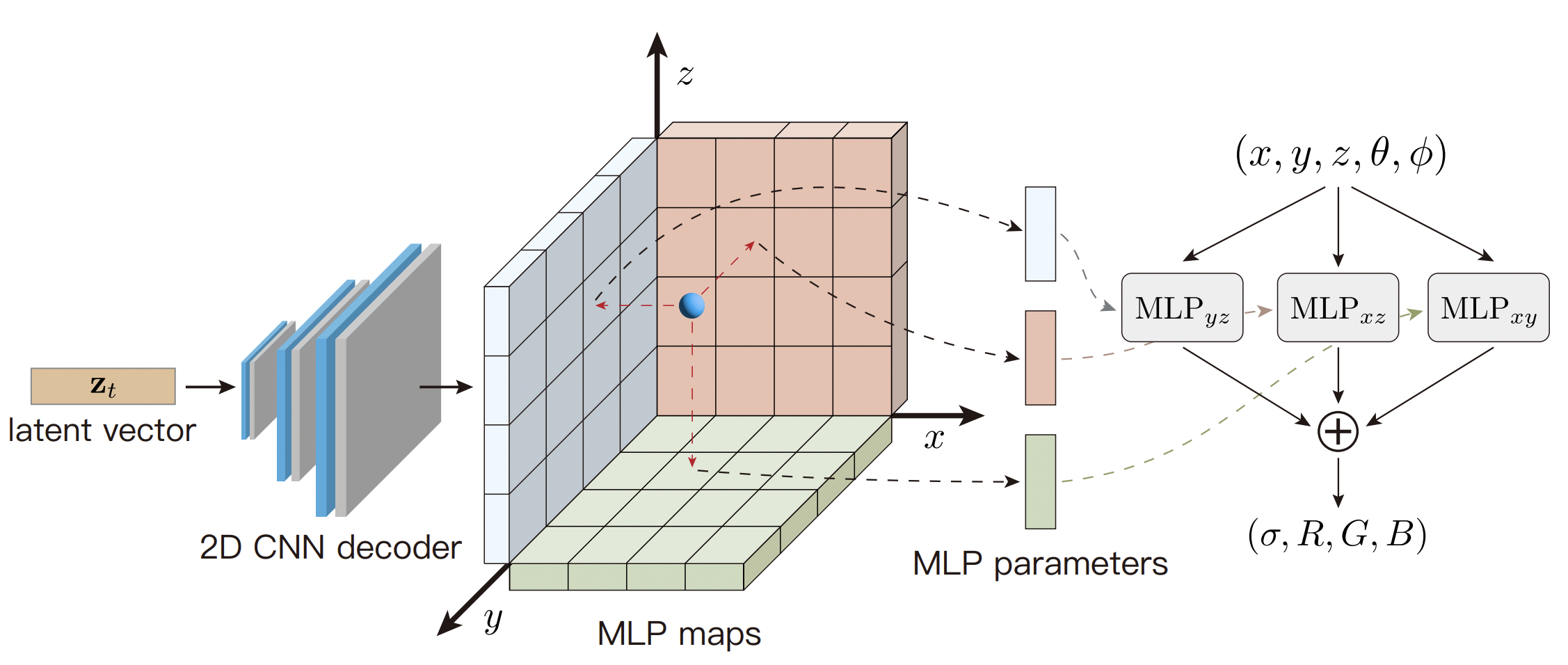} 
    \caption{Structure of a multi-plane method for dynamic NeRF. In this method, a 2D convolutional neural network generates dynamic 2D MLP maps for each video frame, with each pixel containing the parameters of a small MLP. This technique efficiently represents volumetric videos using multiple small MLP networks and improves rendering speed. Image is taken from \cite{peng2023representing}.}
    \label{fig:nerf_multi_plane}
\end{figure}

\subsubsection{Multi-plane and Feature Grids Methods}
\label{nerf_Multi-plane and Feature Grids Methods}
A group of methods \cite{fridovich2023k, cao2023hexplane, hu2023tri, shao2023tensor4d, peng2023representing,icsik2023humanrf, guo2024compact} have proposed representing scenes using feature grids. The core idea involves defining multiple planes, each containing a feature grid, and projecting a point onto these planes to interpolate the features in each grid. The retrieved features from each plane are then combined in order to form a final feature vector, which is subsequently used to generate the final scene. These methods benefit from using compact grid features and a lightweight MLP model, offering better storage efficiency and rendering speed.

Various strategies have been proposed to decompose the 4D spacetime plane into sub-planes. In number of methods \cite{fridovich2023k, cao2023hexplane, hu2023tri, shao2023tensor4d, guo2024compact}, 
the 4D spacetime plane, which includes $X$, $Y$, and $Z$ as spatial parameters along with $T$ as the time dimension, is decomposed into six planes: three spatial planes ($XY$, $XZ$, and $YZ$) and three time-dependent planes ($XT$, $YT$, and $ZT$). Each plane contains a feature grid, a form of explicit representation, that retrieves corresponding parameters by projecting a point onto them. The final features, obtained by interpolating each feature grid, are then used by a lightweight MLP model to quickly predict the final color.
Peng et al. \cite{peng2023representing} proposed using three 2D MLP maps for the spatial dimensions $XY$, $XZ$, and $YZ$ in each video frame. Each cell in these grids contains parameters for a light MLP model that can generate a portion of the final scene. When combined, they produce the entire scene faster. I{\c{s}}{\i}k et al. \cite{icsik2023humanrf} decomposed 4D feature grid into four 3D and four 1D feature grids combined with lightweight MLP models for fast rendering of extensive multi-view data. Figure \ref{fig:nerf_multi_plane} illustrates an example of the basic idea behind methods in this category.


\subsubsection{Rendering Acceleration Methods}
\label{nerf_Rendering Acceleration Methods}
While the previously mentioned methods propose faster representation and rendering algorithms that maintain acceptable storage consumption, another line of research has proposed approaches to speed up NeRF rendering by optimizing the rendering process itself \cite{lombardi2021mixture,garbin2021fastnerf, wang2022fourier,li2023steernerf}. These methods generally aim to reduce the time required for querying the MLP model by utilizing more internal components of the rendering process, rather than adding additional elements such as deformation models or explicit representations like feature grids.

The total computation required to render a NeRF model and produce a 2D image with a resolution of $H \times W$ is given by $H \times W \times (N \times F + C)$, where $N$ represents the number of samples collected along each ray, $F$ denotes the computational cost of querying the MLP model to obtain color and density for each input, and $C$ is a constant associated with integrating the predicted color and density for collected samples to compute the final color for each pixel.

Numerous methods \cite{chen2024nerfhub, zhao2023tinynerf, yuan2024slimmerf, peng2024efficient, reiser2021kilonerf} have been proposed to replace the primary MLP models with more shallow and lightweight models. This strategy helps to achieve faster rendering and reduced memory usage. 

Some methods~\cite{li2023nerfacc, lindell2021autoint, kurz2022adanerf} proposed techniques to reduce the number of required samples to accelerate rendering process. 
Liu et al. \cite{liu2020neural} proposed a new rendering technique called Neural Sparse Voxel Fields which detects space feature voxel grids in the scene. This approach enables NeRF model to skip ray tracing for empty regions of the scene and consequently accelerate rendering.

Although these methods were proposed for static scenes, many have the potential to be applied to dynamic scenes or utilized alongside techniques discussed in other sections to address limitations in the implicit representation of dynamic scenes. 

\subsection{3D Gaussian Splatting}
\label{methods_3dgs}
Despite the remarkable advantages NeRF-based representations offer in terms of visual quality and data storage, their time-consuming training process and computationally intensive rendering have impacted the overall performance of these methods in real-time volumetric video streaming systems. 3DGS, an explicit learnable representation of volumetric content, has been introduced to provide faster training and rendering speeds, facilitating the development of real-time volumetric content delivery.

Similar to NeRF, 3DGS was initially developed for static scenes \cite{kerbl20233d}, and its usage has since expanded to dynamic scenes as well. Like other explicit representations, 3DGS stores information in a data structure that is easy to access, render, and edit. This structure is a point cloud containing many 3D Gaussians that represent a 3D scene. However, creating a separate point cloud for each frame to represent an entire volumetric video is not storage-efficient. To address this problem, researchers have developed methods to take advantage of the fast training and rendering capabilities of 3DGS while mitigating storage and transmission issues. Methods proposed for dynamic 3DGS are generally divided into motion tracking methods (\ref{3dgs_Motion Tracking Methods}) and deformation-based methods (\ref{3dgs_Deformation-based Methods}).


\begin{figure}[h]
    \centering
    \includegraphics[width=0.47\textwidth]{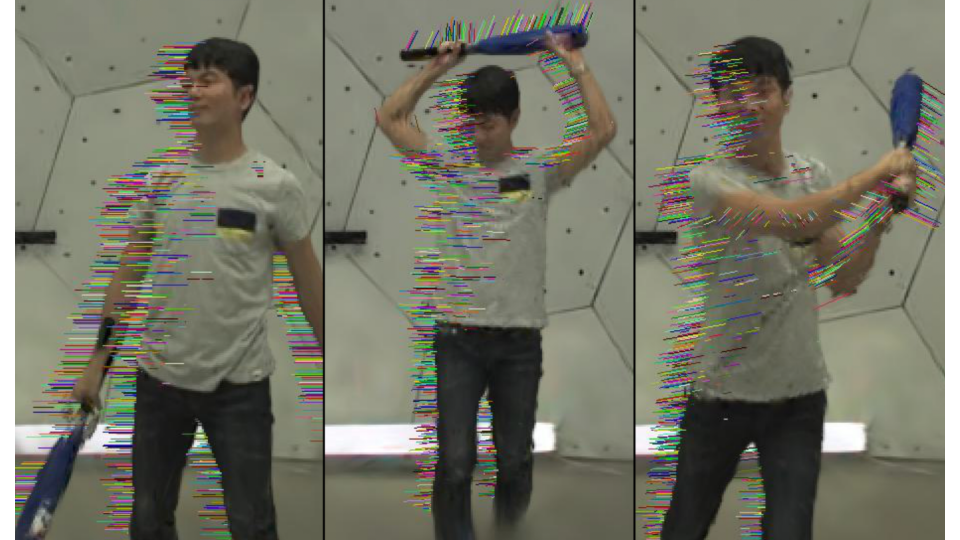} 
    \caption{Results from \cite{luiten2023dynamic}, which introduced one of the earliest motion-tracking methods for dynamic 3DGS.}
    \label{fig:3dgs_motion_tracking}
\end{figure}

\subsubsection{Motion Tracking Methods }
\label{3dgs_Motion Tracking Methods}

In early works on dynamic 3DGS \cite{luiten2023dynamic, sun20243dgstream, katsumata2023efficient, kratimenos2025dynmf, li2024spacetime}, motion-tracking techniques have been used to support the representation of dynamic scenes. These methods typically keep certain attributes static and focus on capturing motion by distinguishing between time-dependent and fixed attributes of the Gaussians. By tracking the motion, they reconstruct each timestep of the volumetric content by modifying only the time-dependent attributes.

In an earlier work, Luiten et al. \cite{luiten2023dynamic} extend the 3DGS representation for dynamic scenes by allowing the position and orientation of Gaussians to change at each timestep, while keeping other attributes such as  color, opacity, and size consistent across all timesteps. Their method uses an online approach to reconstruct frames over time, initializing all Gaussian attributes in the first frame and holding them static for subsequent frames, except for position and orientation, which reflect Gaussians movements. Physical priors including local rigidity, local rotational-similarity, and long-term local-isometry are considered to model each Gaussian motion consistently over time. Figure \ref{fig:3dgs_motion_tracking} illustrates several examples of how this method performs.

Sun et al. ~\cite{sun20243dgstream} propose a similar approach but, rather than optimizing each frame independently, they model Gaussians motion using a Neural Transformation Cache (NTC)~\cite{muller2022instant}. Their method consists of two training phases: in the first phase, an NTC is trained to model Gaussians transformations over time; in the second phase, an adaptive addition strategy is introduced to manage emerging objects.

Similarly, Katsumata et al. \cite{katsumata2023efficient} followed the same idea to consider fixed and time-dependent attributes for Gaussinas to facilitate modeling the motion between frames. they use Fourier approximation to model the position and linear approximation for the rotation.

Kratimenoset al. \cite{kratimenos2025dynmf} consider each scene as a limited set of fixed trajectories and propose efficient basis functions to capture spatial motion. Building on these functions, their method employs motion coefficients to predict dynamic positions and rotations, incorporating constraints like sparsity and rigidity to improve optimization.

Li et al. \cite{li2024spacetime} introduce "Spacetime Gaussians" (STG), which considers time-dependent opacity along with polynomially parameterized motion and rotation to effectively model dynamic content. To minimize model size, spherical harmonic coefficients are replaced by compact features that encode color, view-related  and time-related information.

\begin{figure*}[h]
    \centering
    \includegraphics[width=0.85\textwidth]{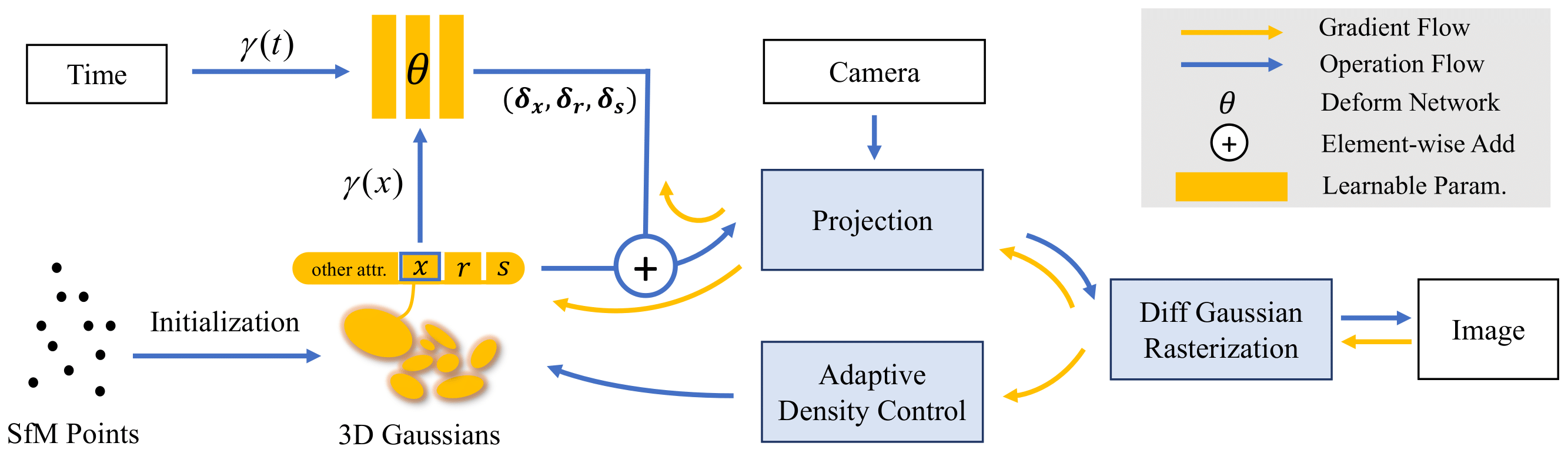} 
    \caption{An example of a deformation-based method pipeline for 3DGS. In this approach, an MLP model predicts offsets ($\delta x, \delta r, \delta s$) for dynamic 3D Gaussians using positional and temporal inputs with positional encoding. The image is taken from \cite{yang2024deformable}.}
    \label{fig:3dgs_deformation}
\end{figure*}

\subsubsection{Deformation-based Methods}
\label{3dgs_Deformation-based Methods}

Another line of work \cite{xiao2024bridging, shaw2023swags, guo2024motion, duisterhof2023md, mihajlovic2025splatfields, wu20244d, lu20243d, liang2023gaufre, yang2024deformable, das2024neural, huang2024sc} 
proposes deformation-based methods, where motion between frames is represented as a deformation field. As explained in Section~\ref{nerf_deformation}, a deformation field is a function that represents the transformations of points (in this case, Gaussians) within dynamic scenes. Methods in this category use a deformation network, which takes a subset of Gaussian attributes and time as inputs to model the deformations of these attributes. The returned deformations can be considered as adjustments to the selected attributes that transform the canonical 3D Gaussians into the deformed space, enabling accurate representation of dynamic scenes.

Most deformation-based methods use an MLP model as the deformation field, which takes the position of Gaussians and time (with positional encoding) as inputs and returns offsets for the position, rotation, and scale of dynamic 3D Gaussians in canonical space. Given a set of 3D Guassians $G(x,s,r,\alpha)$ where $x$ is the central position, $\alpha$ is the opacity and quaternion $r$ and scaling factor $s$ which together make covariance matrix $\Sigma$, prediction the deformation can be modeled as follows:
 \begin{equation}
\delta x, \delta r, \delta s = F_{\text{deform}}(\gamma(x), \gamma(t)),
\end{equation}
where $\delta x$, $\delta r$ and $\delta s$ are offsets for position rotation, and scale of 3D Guassians predicted by the deformation model $F_{\text{deform}}$, and $\gamma(.)$ denotes the positional encoding.

Subsequently, by incorporating the predicted offsets into the 3D Gaussians $G(x+\delta x,s+\delta s,r+\delta r,\alpha)$ the deformation of 3D Gaussians can be modeled over time and the dynamic scene can be represented using 3DGS representation. Figure \ref{fig:3dgs_deformation} the pipeline of a deformation-based method.

Shaw et al. \cite{shaw2023swags} introduce a sliding window training strategy that divides video into smaller windows for handling long-term scenes. An adaptive sampling strategy adjusts window size based on scene motion, balancing training time and visual quality. Then, the optimization process is guided to focus on the dynamic area using an MLP model. In this approach, temporal consistency is maintained through a fine-tuning step using a self-supervised consistency loss on randomly sampled novel frames.

In \cite{yang2024deformable, liang2023gaufre, lu20243d} the static 3DGS is reconstructed in the canonical space first and a deformation model is used to predict the offsets for the position, rotation, and scale attributes.

Das et al. \cite{das2024neural} tackle the challenge of reconstructing dynamic objects from monocular videos. They propose a two-stage approach called Neural Parametric Gaussians (NPGs). In the first stage, they fit a low-rank neural deformation model to preserve consistency. Then, in the second stage, they optimize 3D Gaussians using the learned model from the first stage as regularization.

Wu et al. \cite{wu20244d} propose 4D Gaussian Splatting (4D-GS) which combines 3D Gaussians with 4D neural voxels. In their method, the motion in the scene is represented by a Gaussian deformation field network which contains a temporal-spatial structure encoder and a light multi-head Gaussian deformation decoder. A simplified voxel encoding method, based on HexPlane\cite{cao2023hexplane}, builds Gaussian features from these 4D voxels, and a small MLP then predicts Gaussian movements for new time points.

Gu et al. \cite{guo2024motion} explore the relationship between 3D Gaussian movements and pixel-level flow, introducing a transient-aware auxiliary deformation module to enhance the efficiency and accuracy of deformation prediction.

Deformation-based methods have shown remarkable results in representing dynamic scenes using 3DGS. However, their performance often decreases when there is significant or sudden movement between frames \cite{duan20244d}. Consequently, a number of recent approaches aim to incorporate time as an additional component within 3D Gaussians, directly optimizing the time-dependent Gaussians during training \cite{duan20244d, yang2023real}. Unlike deformation-based or methods based on motion-tracking techniques, this approach does not treat deformation or movement as a separate component and instead, it integrates time directly into the representation to create a unified model for the entire video.

\section{Discussion}
\label{sec:discussion}

Since the introduction of volumetric content, researchers in computer vision, graphics, and systems communities have worked to address key challenges, such as the required computation resources, the high bandwidth demands for transmission on current networks, and the complexities involved in capturing volumetric content. 
Although significant progress has been made through the use of optimized representations and AI techniques, adopting volumetric content as a standard format in media and communication applications remains challenging. 
This section discusses a few promising research directions that may drive the adoption of volumetric content in the near future. 

In general, AI techniques aimed at facilitating volumetric content streaming by targeting several key areas including: (1) reducing the amount of data needed for transmission to the end user, (2) enabling efficient rendering of neural networks that represent 3D scenes by optimizing network structure or using lightweight models that leverage certain features as explicit data, and (3) modeling motion within scenes through motion-tracking and deformation-based methods.

The development of AI techniques has enabled real-time streaming of volumetric content with high visual quality across various datasets. However, deploying these techniques in real-world scenarios presents additional challenges that still need to be addressed. A common limitation in most state-of-the-art methods is their difficulty in capturing large and sudden motions, which hinders the adoption of volumetric video as a primary format in today’s media applications. Developing techniques to model large and sudden motions within frames is an important research direction for future work to address. Another challenge lies in the computational demands on edge devices, though rapid advancements in hardware are expected to bring improvements in this area. Finally, as several AI techniques summarized in this paper have been tested primarily on short video datasets or have shown limitations when applied to long videos, addressing the challenge of streaming long videos remains essential for making volumetric video a practical 3D format for everyday use.

\section{Conclusion}
\label{sec:conclusion}
In this work, we provide a comprehensive overview of recent advancements in the field of volumetric content streaming. 
We first presented a taxonomy of volumetric content representations and then reviewed the primary AI techniques researchers have developed to address the specific challenges associated with each representation. 
At last, we discussed a few promising research directions to address the remaining obstacles in adopting volumetric content streaming. The review will inspire further research that ultimately moving volumetric content streaming closer to becoming a feasible and widespread 3D data format for immersive applications such as entertainment, education, healthcare and communication.

{\small
\bibliographystyle{ieee_fullname}
\bibliography{egbib}
}

\end{document}